\documentclass[1p]{elsarticle}

\usepackage{lineno,hyperref}
\modulolinenumbers[5]

\usepackage{amsmath}
\usepackage{amssymb}

\usepackage{multirow}
\usepackage{subfig}

\journal{Information Fusion}

\begin{document}

\begin{frontmatter}

\title{Combining time-series and textual data for taxi demand prediction in event areas: a deep learning approach\footnote{DOI: https://doi.org/10.1016/j.inffus.2018.07.007\\
URL: https://www.sciencedirect.com/science/article/pii/S1566253517308175}}


\author[mymainaddress]{Filipe Rodrigues\corref{mycorrespondingauthor}}
\ead[url]{http://fprodrigues.com}

\cortext[mycorrespondingauthor]{Corresponding author}
\ead{rodr@dtu.dk}

\author[mymainaddress]{Ioulia Markou}
\ead{markou@dtu.dk}

\author[mymainaddress]{Francisco C. Pereira}
\ead{camara@dtu.dk}

\address[mymainaddress]{Technical University of Denmark (DTU), Bygning 116B, 2800 Kgs. Lyngby, Denmark}

\begin{abstract}
Accurate time-series forecasting is vital for numerous areas of application such as transportation, energy, finance, economics, etc. However, while modern techniques are able to explore large sets of temporal data to build forecasting models, they typically neglect valuable information that is often available under the form of unstructured text. Although this data is in a radically different format, it often contains contextual explanations for many of the patterns that are observed in the temporal data. In this paper, we propose two deep learning architectures that leverage word embeddings, convolutional layers and attention mechanisms for combining text information with time-series data. We apply these approaches for the problem of taxi demand forecasting in event areas. Using publicly available taxi data from New York, we empirically show that by fusing these two complementary cross-modal sources of information, the proposed models are able to significantly reduce the error in the forecasts. 
\end{abstract}

\begin{keyword}
Deep learning \sep Data fusion \sep Cross modality learning \sep Time series forecasting \sep Textual data \sep Taxi demand \sep Special events \sep Urban mobility
\end{keyword}

\end{frontmatter}


\section{Introduction}

Understanding what drives the travel behavior of people is a key research topic for developing effective and efficient intelligent transportation systems that adapt to the travel demand. However, typical approaches focus only on capturing recurrent mobility trends that relate to habitual/routine behaviour \cite{krygsman2004}, and on exploiting short-term correlations with recent observation patterns \cite{moreira2013predicting,Van2015}. While this type of approaches can be successful for long-term planning applications or for modeling demand in non-eventful areas such as residential neighborhoods, in lively and highly dynamic areas that are prone to the occurrence of multiple special events, such as music concerts, sports games, festivals, parades and protests, these approaches fail to accurately model mobility demand \cite{pereira2015using}. Moreover, such scenarios are also known to involve demand surges, which emphasizes the need of good anticipatory capabilities. As we move towards the deployment of autonomous vehicles, understanding and being able to anticipate mobility demand becomes crucial, especially in shared-mobility scenarios, as this allows for properly managing fleets and increasing user-satisfaction. 

In order to capture the effects of events, one can exploit the vast amount of information that is shared online about what is planned to take place in the city. However, most of this information is typically in the form of unstructured natural-language text. Solving this cross-domain data fusion challenge then becomes key for understanding the mobility demand patterns that are caused by events, and also for addressing the general class of problems where text data from the Web can provide the context for explaining some of the patterns that are observed in time-series data. These, not only are quite ubiquitous and cover various research fields, but they are becoming increasingly relevant as people share more and more information online. Popular examples include the use of text data from online social media to help predict opinion polls \cite{o2010tweets} and financial time-series (\mbox{e.g.} stock markets \cite{tang2009stock,si2013exploiting}). 

This paper aims at exploring deep learning architectures for combining time-series and textual data for mobility demand prediction in eventful areas. Specifically, we focus on the problem of taxi demand prediction, although the proposed methodology is applicable to other problems that go beyond the transportation domain. Over the last decade, deep learning has made major advances in solving artificial intelligence problems in different domains such as speech recognition, visual object recognition, object detection and machine translation \cite{schmidhuber2015deep}. It is precisely this success of deep learning in handling different types of data, such as images, audio and text from different domains, that makes it particularly well-suited for the data fusion problem of combining time-series and textual data. Indeed, deep learning has already been shown to be able to outperform traditional approaches for taxi demand forecasting and achieve new state-of-the-art results \cite{zhu2017deep,xu2017real}. However, none of the previous approaches explore Web data about events, particularly in the form of unstructured text, in order to develop more accurate demand forecasting models. This paper aims at bridging this gap. Our proposed neural network architectures combine word embeddings, convolutional layers and an attention mechanism in order to build a joint model of event text descriptions and historical taxi demand data for future demand forecasting. Moreover, the two proposed architectures are quite general and can be applied to similar data-fusion problems from different domains. 

Using a large-scale public dataset of 1.1 billion taxi trips from New York \cite{nyctaxidata} and event data from the Web, we empirically show the value of modelling textual information associated with the events, and that the proposed deep learning approaches are able to outperform other methods from the state of the art by combining information from different sources and formats. Furthermore, we make the source code and datasets used in our experiments publicly available, thus setting them as a data fusion benchmark, so that other researchers can build upon our proposed methodology and use it as a baseline for developing other data fusion methods for the domain application considered in this work. 

The remainder of this paper is organized as follows. In the next section, we review the relevant literature. Section~\ref{sec:model} presents the proposed neural network architectures and the experimental results are presented in Section~\ref{sec:experiments}. The paper ends with the conclusions (Section~\ref{sec:conclusion}).

\section{Related work}
\label{sec:literature}

\subsection{Urban mobility and special events}

Urban mobility demand forecasting under special events has long been recognized to be more difficult than under habitual or recurrent conditions \cite{PotierEtAl03}. However, as our cities grow and become more dynamic, understanding the effect of events in urban mobility becomes crucial for the development of reliable intelligent transport systems. Indeed, for large-scale events (\mbox{e.g.} World cup, Formula 1 and Olympic games), best practices are already available for authorities to follow in order to manage these events and prepare for them well in advance \cite{FHWA06,CoutroubasEtAl03}. Unfortunately, these manual approaches do not scale to the vast amount of smaller and medium-sized events that take place on large metropolitan areas on a daily basis. Despite their reduced scale, these events still have a significant impact in the transportation system \cite{pereira2015so}, especially when multiple co-occur. In these scenarios, common practice relies on reactive approaches rather than on planning \cite{Bolte2006,Middleham2006}. Scalable demand prediction solutions like the ones proposed in this paper, that are based on event data that is automatically mined from the Web, then present themselves with the potential for anticipating the effects of events and proving reliable forecasts for eventful and dynamic areas. 

Due to the importance of special events' impact in urban mobility, its is not surprising that they are a predominant part 
of transportation research. For example, Kuppam et al \cite{Kuppam2011} and Chang and Lu \cite{Chang2013} developed 4-step models where, based on survey response data, they predict, for each event, the number of trips by type, trip time-of-day, trip origin/destination (OD), mode and vehicle miles travelled/transit boardings generated due to the events. However, these works do not consider event characteristics, which are well known to influence the impacts on mobility \cite{CalabresePervasive2010,pereira2013JITS}. For example, in \cite{pereira2013JITS} the authors empirically show that, when available, event categories are indeed valuable inputs for predicting public transport demand. The key difficulty in generalizing these approaches, is that most of the relevant event information is typically available in the form of unstructured text, making it essential to develop cross-domain data fusion techniques such as the ones proposed in this paper, in order to incorporate this information in demand prediction models. 

\subsection{Data fusion of time-series and text}

From a data fusion perspective, combining time-series data with textual information for better understanding real-world phenomena is a very important, yet challenging, problem. The key intuition is that the textual information could contain clues that correlate with the time-series observations and, at least to some extent, explain their behavior. Given the generality of this cross-domain data fusion problem, it is not surprising that different solutions arise from multiple research areas. A particularly prolific one is finance, where Ruiz et al \cite{ruiz2012correlating} analyzed the correlation between micro-blogging activity in Twitter, such as the number of tweets and re-tweets that relate to a company, and stock-market events. Their results suggest that these indicators can be useful to improve trading strategies in the stock market.

In order to improve the predictions of stock prices on the Chinese stock market, Tang et al \cite{tang2009stock} developed an approach that analyses the presence of certain words in news reports that are automatically extracted from the Web. To select the relevant words, the authors rely on a starting set of manually predefined words that are used to identify relevant documents, which then allow them to select the top most relevant words based on their probability ratio (PR). Similarly, Wang et al \cite{wang2012novel} proposed a model that combines ARIMA and support vector regression (SVR) for predicting the quarterly returns of equity (ROEs) of six security companies. The idea is that the ARIMA model is able to analyze the linear autoregressive component of the series, but is complemented by a non-linear SVR model based only on textual feature vectors. In this case, the authors rely on TF-IDF in order to represent the textual data. 

On a different topic, O'Connor et al \cite{o2010tweets} looked at the relationship between text sentiment in tweets and time-series data of public opinion polls. Using a dataset of 1 billion tweets, the authors used a predefined set of keywords to identify tweets related to certain topics, such as the Obama vs McCain election. These were then analyzed according to the sentiment of the words that they contained based on a word-sentiment dictionary. The obtained results show that the frequency of words associated with certain sentiments can be strongly correlated (up to 80\%) with public opinion polls, thus making Twitter, and social media in general, potential alternatives and supplements for traditional polling. On a related work, Tumasjan et al \cite{tumasjan2011election} argues that Twitter is used extensively for political deliberation and that the mere number of party mentions accurately reflects the election result. 

All these works relate to the work presented in this paper by exploring information in textual data in order to explain a time-series of observations. However, they are remarkably different by relying on predefined lexicons of individual words that relate to certain topics or sentiments. The approaches proposed in this paper use convolutional neural network layers to automatically learn word patterns or sequences that relate to the observed time-series data in a joint modeling approach, thereby avoiding the use of ad-hoc heuristics such as sentiment word frequencies and predefined lexicons. 

In the particular domain of urban computing, Pereira et al. \cite{pereira2015using} studied the problem of using event data to help predict public transport demand. Their approach consists of using a bag-of-words representation of event descriptions to learn a topic model, namely latent Dirichlet allocation (LDA), and using the topic assignments as features in a shallow neural network model. The approaches proposed in this paper have the advantage that they do not rely on bag-of-words representation and thus do not ignore the ordering of the words. This makes them capable of capturing multi-word patterns in the text. 

Data fusion approaches are often divided into three main categories \cite{zheng2015methodologies}: (i) approaches that treat different data sources equally and put together the features extracted from them into one single feature vector, (ii) approaches that use different sources of data at different stages and (iii) approaches that feed different datasets into different parts of a model simultaneously. Like a wide majority of the literature, all the approaches described above fall within the first two categories. In contrast, our proposed deep learning models belong to category (iii). 

\subsection{Deep learning in transportation}

Over the last decade, deep learning has proven to be very successful in various applications from different research domains. The fields of transportation and urban mobility are no exceptions. For example, Lv et al. \cite{lv2015traffic} propose to use a stacked auto-encoder (SAE) model to learn generic traffic flow features, that are then used for traffic flow prediction. Based on data from 15.000 individual flow detectors, which are deployed statewide in freeway systems across California, their empirical results suggest that SAEs outperform more traditional approaches based on SVR and radial basis functions (RBFs). Similarly, Ma et al \cite{ma2015long} study the use of long-short term memory (LSTM) networks for travel speed prediction. Their empirical results on data from Beijing indicate that LSTMs outperform other methods such ARIMA and SVR, which the authors justify with the ability of LSTMs to capture long-term dependencies over the time-series. 

On a related work, Zhang et al \cite{zhang2017deep} propose a deep learning architecture based on convolutional layers with residual connections for jointly predicting city-wide crowd flows, \mbox{i.e.} inflows and outflows between regions. Using taxi data from Beijing and bike sharing data from NYC, the authors empirically show that their proposed architecture outperforms standard approaches such as ARIMA and vector auto-regressive models. Lastly, Xu et al \cite{xu2017real} also propose a combination of LSTMs and mixture density networks (MDNs) to predict taxi demand in NYC. In their approach, the city is previously divided in smaller areas and then the LSTM-based model is used to jointly predict the taxi demand for the next time-step (20-60 minutes) in all the areas. 

While the approaches described above demonstrate the potential of deep learning for transportation problems, none of these approaches consider the effect of events in order to improve their predictions. The deep learning approaches proposed in this paper aim at bridging this gap by focusing on event areas and showing that data fusion techniques that combine text data about events and time-series observations of mobility demand can significantly improve predictions. 

\section{Proposed methodology}
\label{sec:model}

In this section, we describe our proposed general-purpose methodology for exploiting contextual text data in time-series forecasting problems using deep learning. We begin by explaining the data preparation techniques used for the time-series and textual data, and then we present two proposed neural network architectures for fusing these two inherently different information sources. 

\subsection{Time-series detrending}

\begin{figure}[t!]
\begin{center}
\includegraphics[scale=0.5]{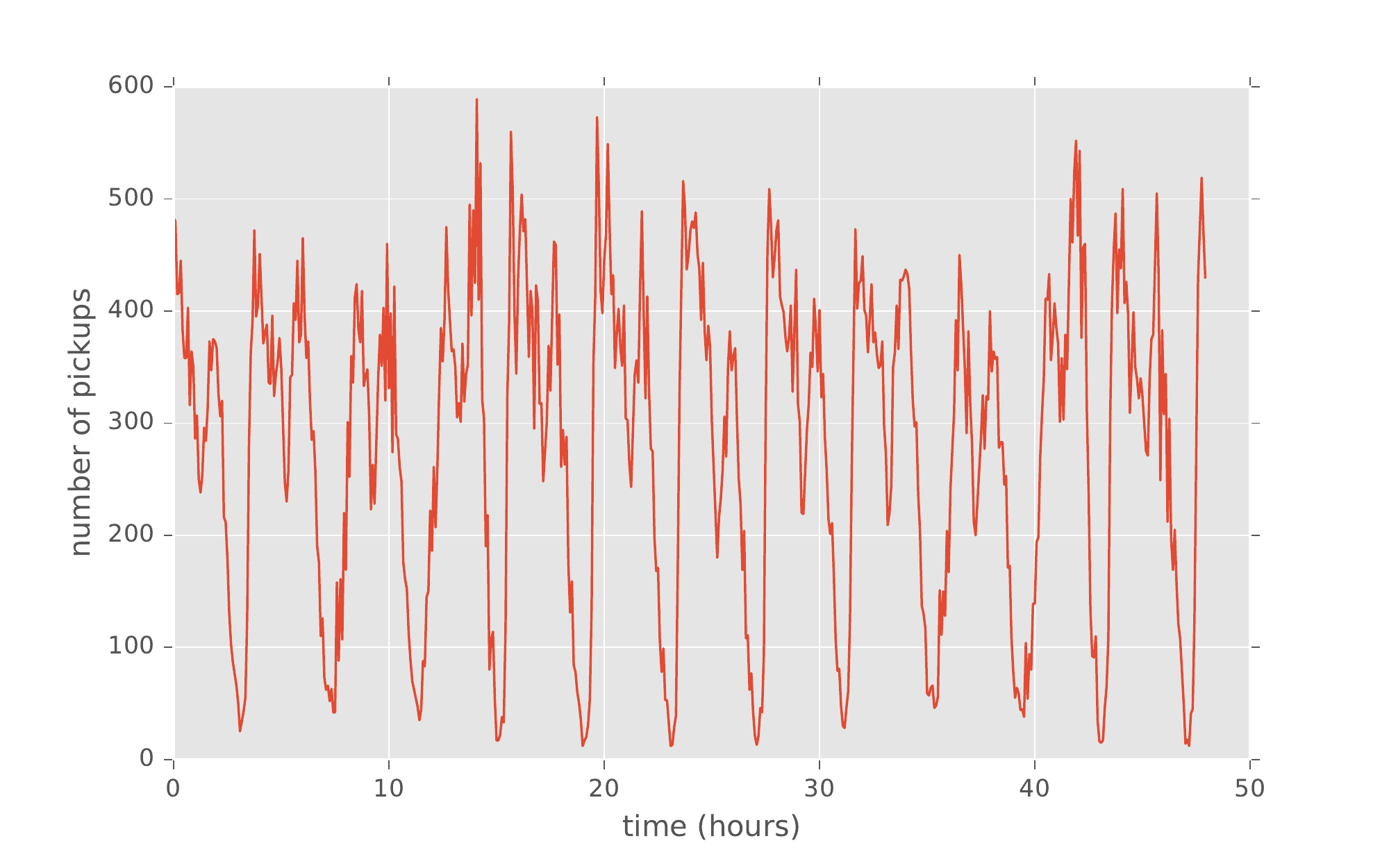}
\caption{Example time-series of pickups in a central area in Manhattan (NYC).}
\label{fig:trend_example}
\end{center}
\end{figure}

One of the most important steps in preparing time-series data for analysis is detrending \cite{wu2007trend}. For the particular case of urban mobility applications, such as traffic flow forecasting and taxi demand prediction, there are obvious cyclic trends that result from daily commuting and other recurrent behaviors. Figure~\ref{fig:trend_example} shows a plot of taxi pickups in an central area of Manhattan (NYC) that illustrates this behavior. A simple, yet very effective way, of identifying these daily or weekly recurring patterns is by constructing a historical averages model, which consists in computing the individual averages for each (time of day, day of the week) pair based on historical data (from the train set only). The historical averages then represent a fixed recurring trend, which can be easily removed from the data. The goal of the prediction models is then to learn to forecast the residuals resultant from the detrending. Based on our experience with various time-series forecasting problems with urban mobility data, by removing the burden of capturing these well-known recurrent trends from the model, this simple detrending procedure significantly improves the predictive performance. In fact, as our experiments demonstrate, it makes even the simplest linear models quite competitive baselines to outperform. 

\subsection{Text data preprocessing}

When working with textual data mined from the Web in supervised learning models, it is typically important to preprocess it in order to make it more amenable for learning methods. In this work, we follow a simple conventional text-preprocessing pipeline consisting of:
\begin{itemize}
\item HTML tags removal - this is particularly important when working with data extracted from the Web;
\item Lowercase transformation - reduces the variability in the way words can appear in the text;
\item Tokenization - divides a sequence of characters into pieces called \textit{tokens};
\item Lemmatization - removes inflectional endings of words and returns the base or dictionary form of a word, which is known as the lemma;
\item Removal of stopwords and very frequent words - removes words such as ``a", ``the" and ``and", which typically do not bring any additional useful information; 
\item Removal of words that appear only once - like very frequent words, these words are often not very informative; in our experiments, we found that removing these words significantly reduces the vocabulary size and, consequently, the neural network complexity, without reducing the predictive performance.
\end{itemize}
Please note that, while the word embeddings used in our proposed neural network architecture (see Section~\ref{subsec:nn}) are expected to handle some of the issues covered by the text preprocessing pipeline (\mbox{e.g.} by representing the words such as ``sport" and ``sports" close to each other in the embedding space), we believe that doing these transformations during the preprocessing stage is still a safer and, therefore, preferable approach. Moreover, it significantly reduces the size of the vocabulary used by the embeddings layer - a computational advantage.

\subsection{Neural network architecture}
\label{subsec:nn}

There are several ways in which one can combine text data with time-series observations using deep learning. In this section, we present two neural network architectures for this problem, which differ essentially on the way the time-series data is modeled. We experimented extensively with multiple variations of these architectures, and the proposed architectures are the ones that consistently performed best. 

\begin{figure}[t!]
\begin{center}
\includegraphics[scale=0.11]{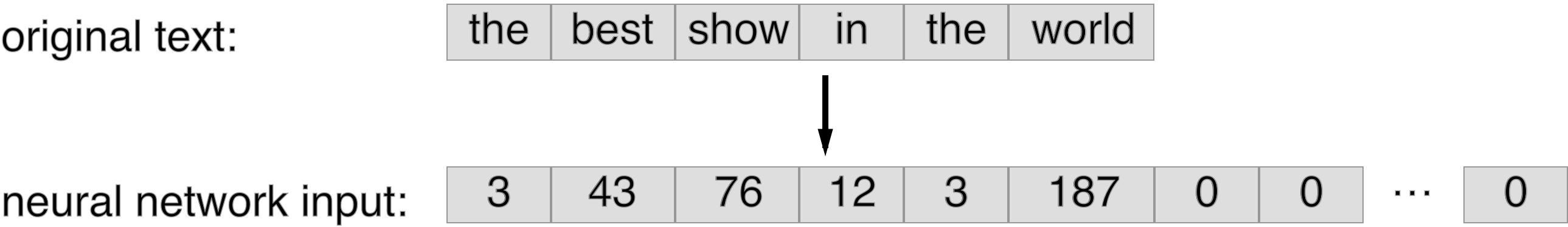}
\caption{Example conversion of text into the vector input of the neural network.}
\label{fig:text_input}
\end{center}
\end{figure}

Common to both proposed architectures is the way textual data is incorporated in the model. It consists of a combination of word embeddings and convolutional layers. The textual input to the network is a 1D vector with each element consisting of an integer identifier of the corresponding word in the text, so that multiple occurrences of a word are given the same identifier. Figure~\ref{fig:text_input} illustrates this process. The text input to the network therefore corresponds to raw text (after the text preprocessing described earlier) in the form of a 1-dimensional vector of words identifiers (integers). The length of the vector corresponds to a predefined maximum length, $S$, that is typically set to the longest input word sequence in the train set. Note that this value is important mainly for computational reasons. The sequences that are shorter than this value are padded with zeros. 

The vector of integers representing a sequence of text then goes into a word embedding layer. The idea of word embeddings is to map semantic meaning into a geometric space by assigning to each word in a dictionary a numeric vector, such that the distance between any two vectors in the embedding space captures the semantic relationship between their corresponding words. For example, the words ``music" and ``rock" are expected to be geometrically close in the embedding space, since they are closely related to each other. On the other hand, the L2-distance in the embedding space between words that are semantically very different, such as ``theater" and ``basketball", should be very large. Good word embeddings further allow for semantically-meaningful operations between word embedding vectors, such as: ``king" + ``woman" = ``queen". 

Word embeddings are typically obtained by applying unsupervised learning techniques to datasets of co-occurrence statistics between words in a very large corpus. In this paper, we use the popular GloVe embeddings \cite{pennington2014glove}, which were obtained by applying a matrix factorization technique to a 2014 dump of the English version of Wikipedia. Namely, based on a matrix of word co-occurrence frequencies in the whole corpus, the authors learn word vectors such that their dot product equals the logarithm of the words' probability of co-occurrence, which is represented as a weighted least-squares objective. The embeddings layer in our proposed deep learning architecture is then initialized with the word embeddings produced by GloVe. We use 300-dimensional embedding vectors for each word. 

\begin{figure}[t!]
\begin{center}
\includegraphics[width=\textwidth,height=\textheight,keepaspectratio]{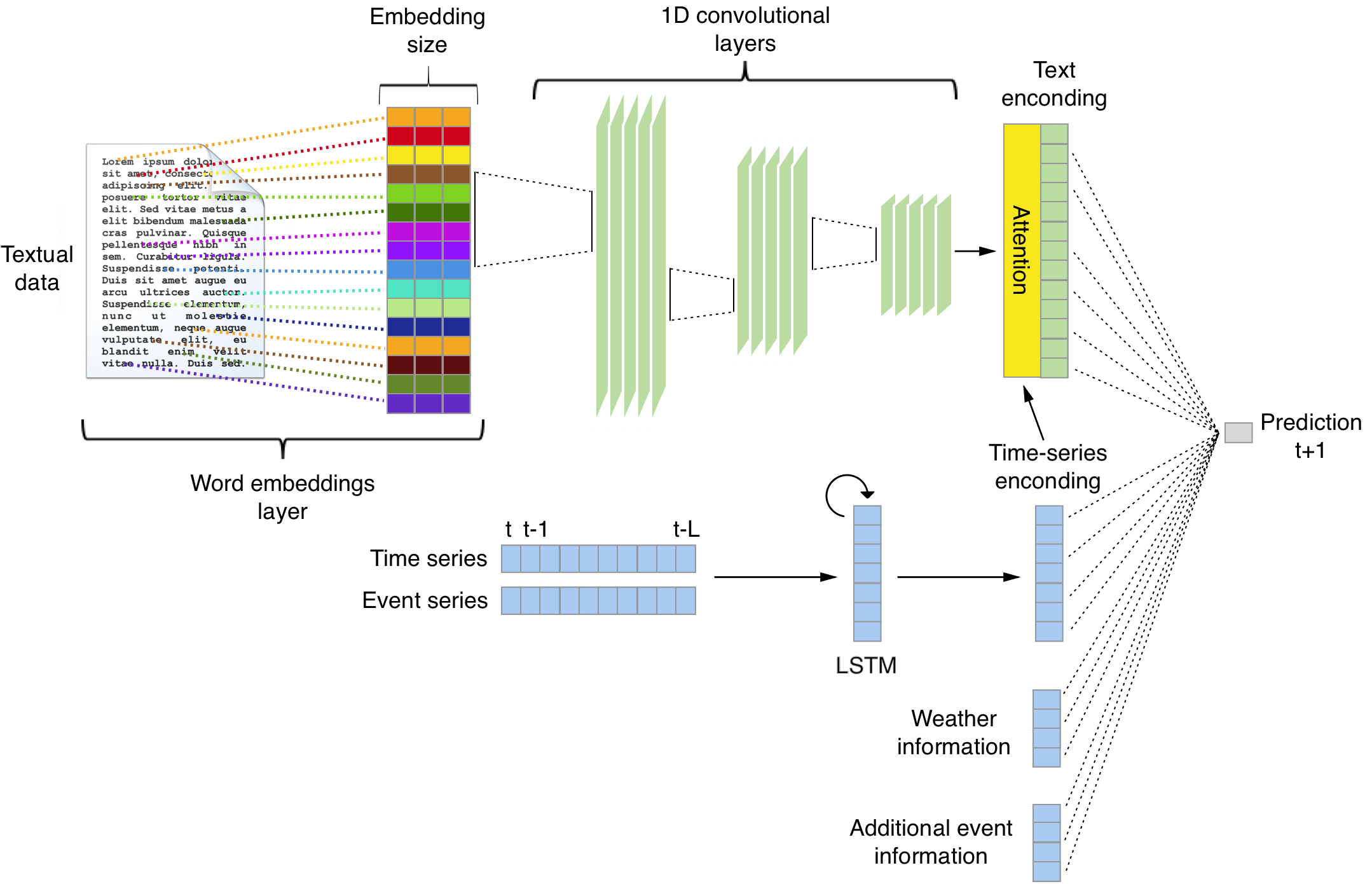}
\caption{Proposed neural network architecture with an LSTM layer for modelling the time-series observations (DL-LSTM).}
\label{fig:architecture_lstm}
\end{center}
\end{figure}

The numeric vectors of the word embeddings in a text are put together in a 3D tensor of size $S\times1\times300$, where $S$ denotes the maximum sequence length. This tensor in then passed to a series of 1D convolution filters and max-pooling layers that are able to learn to detect certain patterns in the text at different levels of abstraction, in a way that is akin to the way convolutional neural networks (CNNs) learn to detect features in images. Notice how the last dimension (300) of the tensor corresponds to the ``depth" or ``number of channels" in the CNN literature. More concretely, we use a sequence of three convolutional layers with sizes $3\times3$, $3\times3$ and $5\times5$, and 50, 30 and 30 filters, respectively. Each convolutional layer uses ReLU activations and is followed by a max-pooling layer with the same size as the preceding convolution. Between the convolutional layers we apply Dropout \cite{srivastava2014dropout} with a 50\% probability of keeping a connection. The output of the last layer goes through an (optional) attention mechanism and is then combined with the network's learned representation of the time-series data. The top part of Figure~\ref{fig:architecture_lstm} illustrates the neural network architecture for modeling the textual data. Due to the fact that the proposed architecture is based on convolutional layers (and due to the nature of convolutions), it is able to capture variable-length patterns in input text sequences of arbitrary sizes. 

With the purpose of putting more emphasis on the most prominent pieces of textual information, a simple soft attention mechanism, inspired by the one described in \cite{cho2015describing}, was developed. This attention mechanism takes as input the text representation at the last convolutional layer, denoted as $\textbf{z}$, and based on the context $\textbf{c}$ provided by the latent representation of the time-series data decides which parts of the text are more relevant for making a prediction for $t+1$. Specifically, the output of the attention layer $\textbf{h}$ is given by the element-wise product $\textbf{h} = \boldsymbol\alpha \odot \textbf{z}$, where the weights $\alpha_i$ are given by a softmax function:
\begin{align}
\alpha_i &= \frac{\exp(e_i)}{\sum_{k=1}^{K} \exp(e_k)} \nonumber
\end{align}
where 
\begin{align}
e_{i} &= a_i(z_i, \textbf{c}) \nonumber
\end{align}
is a learned relevancy function that scores the extracted text feature $z_i$ given the context vector $\textbf{c}$ and is defined as a fully-connected layer with an hyperbolic tangent (tanh) activation.  

In order to model the time-series data, we consider two different approaches: one based on recurrent neural networks (RNNs), namely LSTMs, and another approach based on fully-connected (FC) dense layers. The motivation for using LSTMs comes from the multiple recent works showing their successful application to time-series forecasting problems, including taxi demand forecasting \cite{ma2015long,xu2017real}. This success is often attributed to their ability to capture long-term dependencies in the data. However, RNNs are known to be significantly harder to train \cite{schmidhuber2015deep}. Moreover, if effective detrending techniques are applied to time-series data of mobility demand such that the clear recurrent daily and weekly patterns are removed, the need for capturing long-term dependencies can be, to a great extent, eliminated. This could potentially make LSTMs unattractive for modeling certain time-series, as it can lead to overfitting. Indeed, some previous research works argue against the use of LSTMs for certain time-series forecasting problems in favor of simpler methods based on FC layers \cite{gers2002applying}. Motived by these issues, we also developed a deep learning architecture that relies on FC layers for modeling the time-series data. 

We begin by describing the data fusion architecture that makes use of LSTMs, which is illustrated in Figure~\ref{fig:architecture_lstm}. As the figure suggests, the time-series data goes into a LSTM layer, together with a vector of binary variables indicating whether the corresponding value at time $t$ in the time-series has or not an event, so that the hidden state of the LSTM cell can also take that information into account for determining the next state. In order to prevent overfitting, we apply regularization to weights of the LSTM. The vector corresponding to the last hidden state of the LSTM cell is then assumed to have encoded the relevant time-series information that is required for making a prediction for $t+1$. 

Let $\textbf{z}_{\mbox{\scriptsize ts}}$ and $\textbf{z}_{\mbox{\scriptsize text}}$ denote the latent representations learned by the proposed neural network architecture for the time-series and textual data, respectively. In other words, $\textbf{z}_{\mbox{\scriptsize ts}}$ is a vector corresponding to the output of the LSTM (or FC) layers that encode the time-series data, while the vector $\textbf{z}_{\mbox{\scriptsize text}}$ corresponds to the output of the attention mechanism and convolutional layers that model the textual data. Notice that these learned representations can be very complex functions of their respective inputs. In the last layer of the network, these two learned representations are combined through a fully connected (dense) layer that computes a final prediction for $t+1$ as the dot product between two vectors of adjustable weights $\textbf{w}_{\mbox{\scriptsize ts}}$ and $\textbf{w}_{\mbox{\scriptsize text}}$ and the corresponding latent representations ($\textbf{z}_{\mbox{\scriptsize ts}}$ and $\textbf{z}_{\mbox{\scriptsize text}}$). In the case that other relevant information is available, such as information about the weather or other details about the events, these can be easily included as input to the last dense layer of the model as Figure~\ref{fig:architecture_lstm} illustrates. We refer to this deep learning approach as ``DL-LSTM", since it uses an LSTM to encode the time-series data.

\begin{figure}[t!]
\begin{center}
\includegraphics[width=\textwidth,height=\textheight,keepaspectratio]{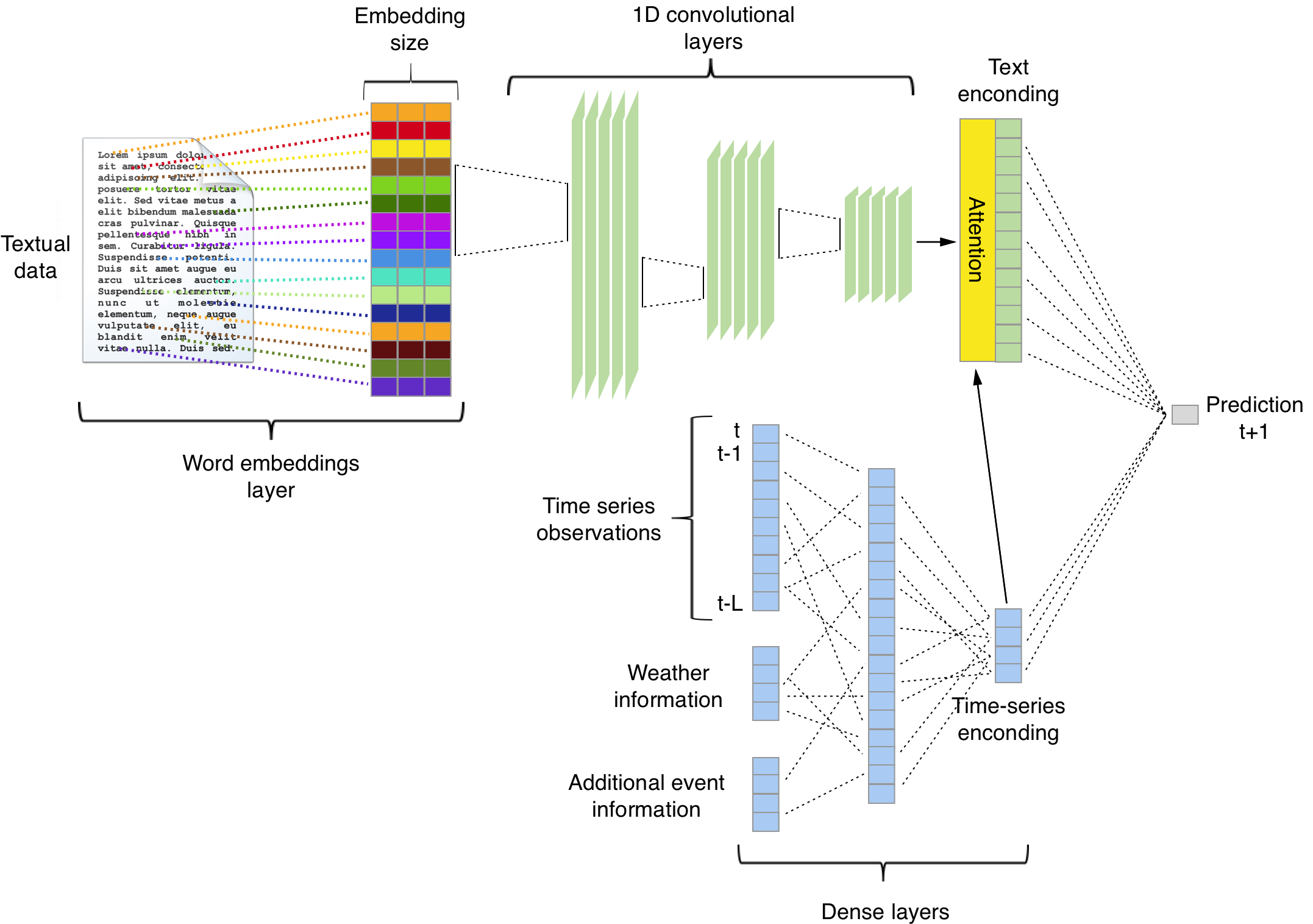}
\caption{Proposed neural network architecture with FC layers for modelling the time-series observations (DL-FC).}
\label{fig:architecture_fc}
\end{center}
\end{figure}

As for the data fusion architecture that makes use of fully-connected (FC) layers for modeling the time-series data, it is depicted in Figure~\ref{fig:architecture_fc}. We refer to this deep learning approach as ``DL-FC" in order to distinguish from the one described above (``DL-LSTM"). In the case of DL-FC, all the time-series information is provided as a flat input vector to the network in the form of ``lagged observations" or \textit{lags}. \mbox{I.e.}, the network is fed with the values for the observations at times $\{t,t-1,\dots,t-L\}$ in a vector of size $L+1$, where $L+1$ corresponds to the number of lags. This vector of lags is fed into a FC layer with 100-200 hidden units and hyperbolic tangent (tanh) activations, which can also receive additional inputs with other relevant information such as weather data or events details. The output of this FC layer is then passed to a second FC layer with 50 units and tanh activations. We apply BatchNormalization \cite{ioffe2015batch} before every FC layer, Dropout between FC layers and we use regularization whenever necessary. The idea is that the output of the last FC layer corresponds to a latent vector representation that encodes all the necessary information from the time-series and other relevant inputs (\mbox{e.g.} weather). This latent vector representation is then combined with the latent representation of the textual data in order to produce a prediction for $t+1$ using a dense layer. The final prediction is obtained by adding back the removed recurrent trend (based on the historical average) to the output of the neural network. 

All layers of the two proposed deep learning architectures for data fusion of time-series with text data are trained via backpropagation using the Adam optimizer \cite{kingma2014adam} and mini-batches of size 64. We use a separate validation set in order to keep the track of the best performing model during training. This validation set was also used to guide the majority of the design and hyper-parameter choices that the deep learning architectures described above reflect. 

\section{Experiments}
\label{sec:experiments}

The proposed data fusion approaches for combining time-series data with textual information were implemented in Keras \cite{chollet2015keras}. Source code with the implementation of the proposed neural network architectures and for reproducing all the results provided in this paper is provided as supplementary material. We proceed by evaluating the proposed approaches in the transportation domain. Namely, we consider the problem of taxi demand forecasting in areas with events. 

\subsection{Dataset and case studies}

\begin{figure}[t!]
\begin{center}
\includegraphics[scale=0.47]{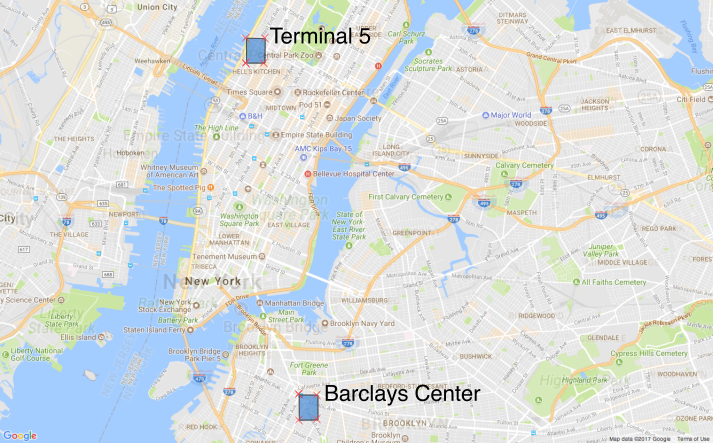}
\caption{Map of the two study areas.}
\label{fig:areas}
\end{center}
\end{figure}

The base dataset for our experiments consists of 1.1 billion taxi trips from New York (January 2009 to June 2016) that were made publicly available by the NYC Taxi \& Limousine Commission \cite{nyctaxidata}. Based on this data, we looked at a list of the top venues in NYC \cite{topvenues} and selected the two venues for which more complete event records where available online: the Barclays Center and Terminal 5. Located in Brooklyn, the Barclays Center is modern multi-purpose arena with 18.000 seats that regularly hosts major musical performances and serves as the new home of the NBA's Brooklyn Nets. On the other hand, the Terminal 5 is a 3-floor venue that regularly hosts concerts with many different audiences and that is located in the heart Manhattan. Given the geographical coordinates of these two venues, we selected all the taxi pickups that took place within a bounding box of $\pm$0.003 decimal degrees (roughly 500 meters) to be our study areas. Figure~\ref{fig:areas} shows a map of these areas. 

The individual pickups were grouped in a time-series of daily counts. The goal is then to predict the number of taxi pickups in an area for the next day given the pickups from the previous days, weather data and event information extracted from the Web. From the perspective of taxi operators and companies like Uber and Lyft, knowing in advance how many taxi pickups are expected for the next day allows them to allocate the required number of drivers well in advance. Similarly, from a perspective of autonomous vehicles and shared mobility services such as DriveNow and Car2Go, this allows providers to re-allocate their fleets during the night according to the predicted demand for the next day. Nevertheless, it is important to recall that the approaches proposed in this paper are quite general and can be applied to other problems that span various domains. 

The weather data was obtained from the National Oceanic and Atmospheric Administration (NOAA) and corresponds to measurements from a weather station located in the Central Park in NYC. It contains information about minimum and maximum temperatures, precipitation, wind, presence of snow, fog, snow depth, etc. 

Regarding the event data, it was extracted automatically from the Web using either screen scrapping techniques or API's. For the Barclays Center, the event information was scrapped from its official website, since it maintains a very accurate and detailed calendar that allowed us to go back in time and retrieve data for the period matching the taxi demand data. We collected a total of 751 events since its inauguration in late 2012 until June 2016. As for the Terminal 5, we used the Facebook API to extract 315 events for a similar time period. In both cases, the event data includes event title, date, time and description. It is important to note that, in practice, one could easily use one of the many event directories and aggregators available online, such as Eventful.com, Timeout.com, etc. Our choice to rely on these specific ones was to facilitate the retrieval of past events, which is typically not available in generic online event directories. 

\begin{figure*}
\centering
\subfloat[Barclays Center]{\includegraphics[width=0.5\linewidth]{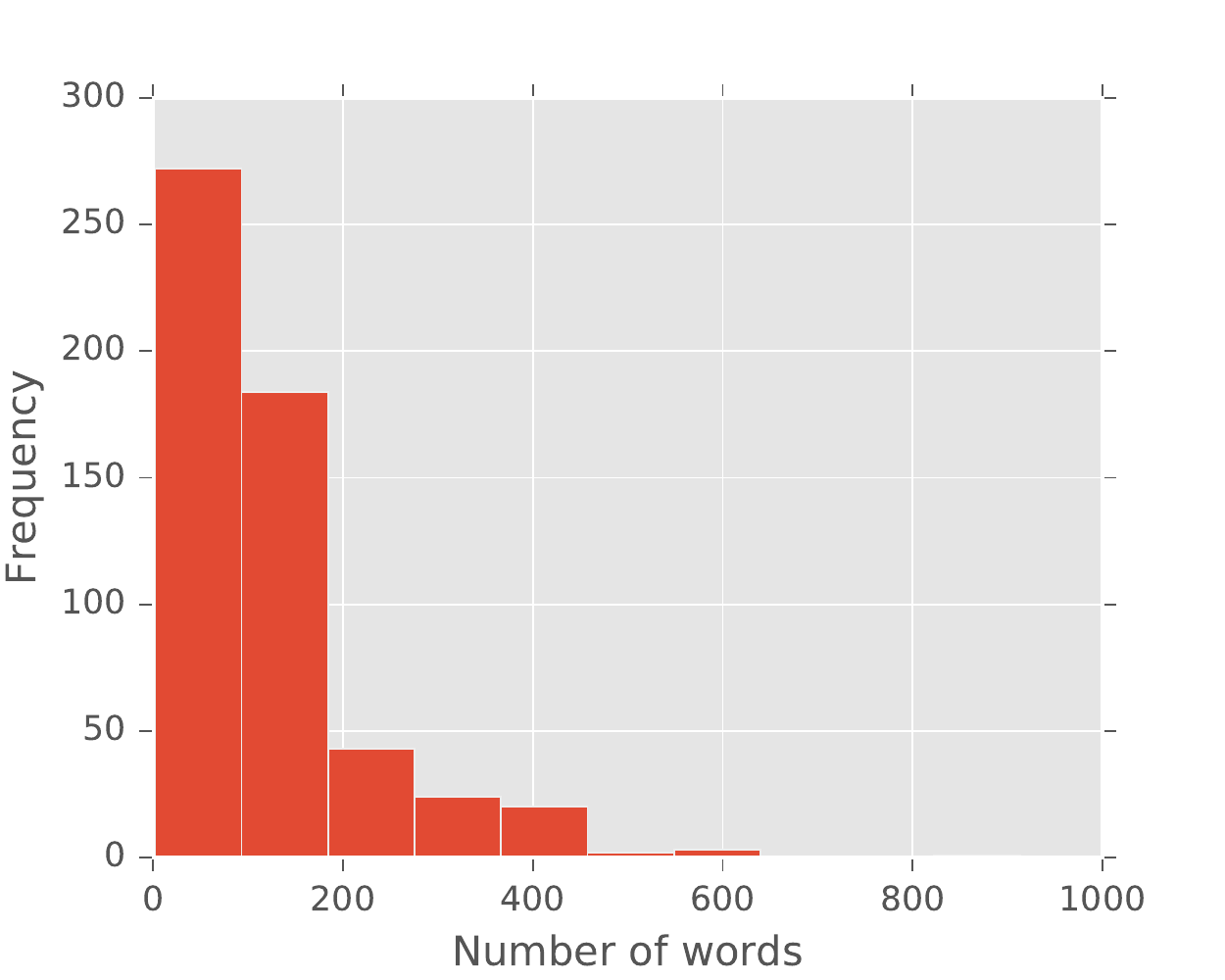}\label{fig:hist_barclays}}\hspace{-0.2cm}
\subfloat[Terminal 5]{\includegraphics[width=0.5\linewidth]{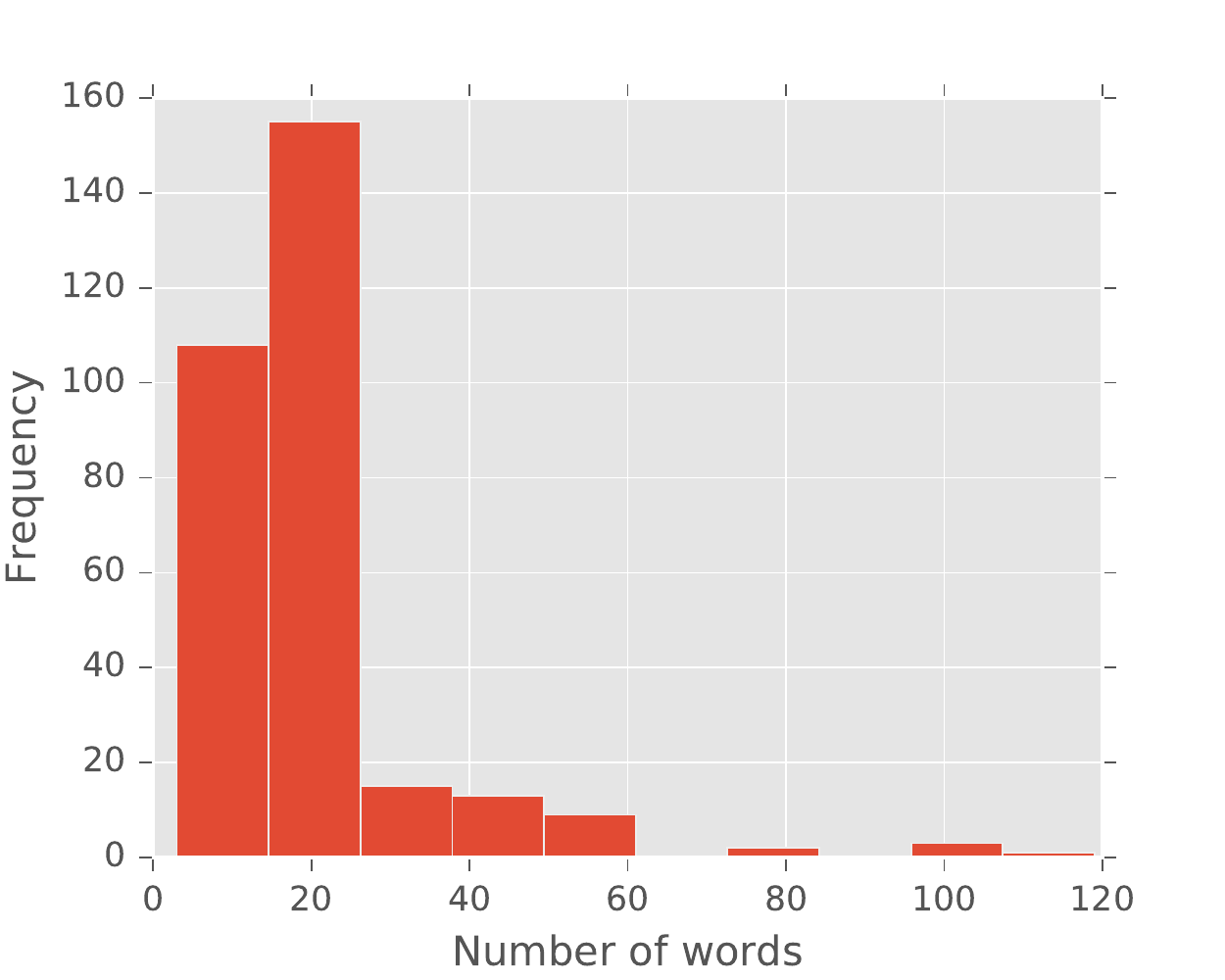}\label{fig:hist_term5}}
\caption{Histograms of words counts in the event descriptions from (a) Barclays Center and (b) Terminal 5}
\label{fig:hist_word_counts}
\end{figure*}

Table~\ref{table:events} shows 5 examples of the event information collected that illustrate the diversity and heterogeneity of this data. In particular, notice how the length and quality of the descriptions can vary. In order to analyse this aspect further, Figure~\ref{fig:hist_word_counts} shows histograms of the number of words for the events from each study area. The average number of words per event description is 113 (std. $\pm 114$) and 20 (std. $\pm 16$) for Barclays Center and Terminal 5, respectively. In fact, the number of words in the text can be higher than 500. However, this is not an issue for the proposed neural network architecture, since it can capture patterns in input text sequences of arbitrary size due to the use of convolutional layers. Lastly, since some of the events occur late at night, they may contribute to a sharp rise in taxi demand during next day's dawn, and thus at next day's aggregate demand. Therefore, we derived additional features in order to account for this effect. 

\begin{table}[t!]
\caption{Five examples of events and descriptions extracted from the Web.}
\begin{center}
\begin{tabular}{p{3cm}|p{8cm}}
Title & Description\\
\hline
Walk the Moon at Terminal 5 on 4/14 & WALK THE MOON at Terminal 5 on 4/14 (Sold Out)  All Ages \\
\hline
Local Natives \& with Charlotte Day Wilson & Charity: Local Natives believe in equality, safety, and dignity of all people. They have partnered with Plus 1 so that \$1 from every ticket is going to support gender-based violence intervention... \\
\hline
Ringling Bros. \mbox{ } Circus	 & Witness the Greatest Show On Earth one last time! Be a part of the last curtain call when Ringling Bros. and Barnum \& Bailey presents Out Of This World, coming to Barclays Center...\\
\hline
Arcade Fire & Due to overwhelming demand, Grammy Award-winning band, Arcade Fire, announced additional dates for the highly-anticipated REFLEKTOR TOUR in support of its international \#1 album...\\
\hline
2014 NBA Playoffs - Nets vs. Heat - Game 3 & Its Game 3 of the second round of the NBA Playoffs, Saturday, May 10 when the Brooklyn Nets take on the Miami Heat at Barclays Center...\\
\end{tabular}
\end{center}
\label{table:events}
\end{table}%

\subsection{Experimental setup}

Based on the data from the two case-study areas described in the previous section, we created separate training, validation and test sets. For the Barclays Center, the training set consists of 21 months of observations (January 2013 to September 2014), the validation set contains the remaining data from 2014 (3 months), and the data from January 2015 to June 2016 (18 months) is used for testing the different approaches. Regarding Terminal 5, since the frequency of events is smaller, we increased the size of validation set in order to contain a representative number of events. The train set then consists of 2 years of data (2013 and 2014), the year of 2015 is used for validation and the remaining data from 2016 is used for testing. We trained independent models for each study area. 

In order to evaluate the contribution of the different sources of information, we perform an incremental analysis of the proposed deep learning architectures where we start with only the part of the network that is responsible for modeling the time-series data and we keep adding components to the network until the full models depicted in Figures~\ref{fig:architecture_lstm} and \ref{fig:architecture_fc} are obtained. Therefore, we start with a model that only takes the lagged observations or \textit{lags} (referred to as ``L") as input, and move to models that also include: weather information (denoted ``L+W"), information about the presence of events (``L+W+E") and, finally, the full model that also considers the event's textual descriptions (``L+W+E+T"). Please note that, in their simplest versions (``L"), the proposed models - DL-LSTM and DL-FC - are respectively equivalent to using plain LSTM and standard feed-forward neural network approaches for time-series forecasting. 

Besides these baselines, the proposed approach is further compared with two popular methods from the state of the art for time-series forecasting: support vector regression (SVR) and Gaussian processes (GPs). The SVR approaches use a linear kernel, since our experimental results showed that it provides the best results for this particular problem. As for the GPs, we use a composite covariance function that consists of the sum of a squared exponential and a white noise covariance function. The hyper-parameters of both the SVR and the GP were tuned based on their performance on the validation set using an exhaustive search procedure of all possible combinations within the set $\{0.001,0.01,0.1,1.0,10,100\}$. For both these baseline methods, GPs and SVR, we also perform the same incremental analysis of the contribution of the different sources of information to the predictive performance that was described above. 

In order to evaluate the different approaches, we compute and report the following error statistics:
\begin{align}
\mbox{MAE} &= \frac{1}{N} \sum_{n=1}^N {|y_n - \hat{y}_n|},\nonumber\\
\mbox{RMSE} &= \sqrt{\frac{1}{N} \sum_{n=1}^N {(y_n - \hat{y}_n)^2}},\nonumber\\
\mbox{MAPE} &= \frac{100}{N} \sum_{n=1}^N \bigg| \frac{y_n - \hat{y}_n}{y_n}\bigg|,\nonumber\\
\mbox{R}^2 &= 1 - \frac{\sum_{n=1}^N {(y_n - \hat{y}_n)^2}}{\sum_{n=1}^N {(y_n - \bar{y})^2}},\nonumber
\end{align}
where $N$ denotes the number of testing instances, $y_n$ and $\hat{y}_n$ correspond to true and predict values for the $n^{\mbox{th}}$ instance, respectively, and $\bar{y}$ denotes the average value of $y$. 

\subsection{Results}

\begin{table}[t!]
\caption{Average results and standard deviations for the Barclays Center area over 30 executions of the different approaches.}
\begin{center}
\small
\setlength\tabcolsep{5.2pt} 
\begin{tabular}{l|c|c|c|c}
Method & MAE & RMSE & MAPE & R$^2$ $(\times 100)$\\
\hline
SVR L & 120.7 ($\pm$0.0) & 167.5 ($\pm$0.0) & 15.4 ($\pm$0.0) & 42.8 ($\pm$0.0)\\
SVR L+W & 120.1 ($\pm$0.0) & 166.4 ($\pm$0.0) & 15.4 ($\pm$0.0) & 43.6 ($\pm$0.0)\\
SVR L+W+E & 97.9 ($\pm$0.0) & 139.1 ($\pm$0.0) & 12.6 ($\pm$0.0) & 60.5 ($\pm$0.0)\\
GP L & 145.2 ($\pm$0.0) & 199.4 ($\pm$0.0) & 16.3 ($\pm$0.0) & 18.9 ($\pm$0.0)\\
GP L+W & 127.3 ($\pm$0.0) & 176.9 ($\pm$0.0) & 15.2 ($\pm$0.0) & 36.2 ($\pm$0.0)\\
GP L+W+E & 100.0 ($\pm$0.0) & 141.1 ($\pm$0.0) & 12.8 ($\pm$0.0) & 59.4 ($\pm$0.0)\\
DL-LSTM L & 129.0 ($\pm$2.6) & 172.6 ($\pm$4.3) & 15.9 ($\pm$0.2) & 39.2 ($\pm$3.1)\\
DL-LSTM L+W & 128.9 ($\pm$1.9) & 167.8 ($\pm$2.0) & 16.5 ($\pm$0.2) & 42.6 ($\pm$1.4)\\
DL-LSTM L+W+E & 103.9 ($\pm$1.2) & 141.5 ($\pm$1.6) & 13.3 ($\pm$0.3) & 59.2 ($\pm$0.9)\\
DL-LSTM L+W+E+T & 100.4 ($\pm$1.8) & 139.5 ($\pm$2.0) & 12.8 ($\pm$0.3) & 60.3 ($\pm$1.2)\\
DL-FC L & 120.8 ($\pm$1.1) & 161.7 ($\pm$0.6) & 15.8 ($\pm$0.3) & 46.7 ($\pm$0.4)\\
DL-FC L+W & 119.8 ($\pm$1.0) & 160.5 ($\pm$0.6) & 15.6 ($\pm$0.2) & 47.5 ($\pm$0.4)\\
DL-FC L+W+E & 95.6 ($\pm$0.9) & 134.8 ($\pm$0.8) & 12.7 ($\pm$0.2) & 63.0 ($\pm$0.4)\\
DL-FC L+W+E+T & \textbf{93.2} ($\pm$0.8) & \textbf{132.3} ($\pm$0.9) & \textbf{12.3} ($\pm$0.1) & \textbf{64.3} ($\pm$0.5)\\
\end{tabular}
\end{center}
\label{table:barclays}
\end{table}%

Table~\ref{table:barclays} shows the obtained average results over 30 executions of the different approaches for the Barclays Center area. There are several very interesting aspects that are worth discussing. Let us start by comparing the performance of the simplest versions (``L") of each prediction method. Interestingly, our results show that LSTMs are actually not the best performing method for this particular time-series forecasting problem. We experimented extensively with different variations and hyper-parameters of the LSTMs and we were unable to match the results of its fully-connected feed-forward counterpart (DL-FC). This supports the findings of earlier works such as \cite{gers2002applying}, who suggest that the superiority of LSTMs does not carry over to certain simpler time-series prediction tasks solvable by time window approaches. Indeed, based on our experimental results in Table~\ref{table:barclays}, SVR and DL-FC are the best performing methods for this study area. 

Let us now analyze the contributions of the different sources of information. From Table~\ref{table:barclays} we can verify that including information about the weather improves the predictions. This is reasonable since factors such as rain or cold temperatures are expected to increase the likelihood of people opting for taxi as their mode of transport. However, the source of information that has the greatest impact in the prediction performance of the different methods is the information about the events. For example, in the case of DL-FC, including this information (``E") leads to a MAE reduction of 20.8\% and 16.0\% in RMSE. Including the textual information about the events (``T") in the DL-FC model allows it to further reduce MAE by another 2.5\%, and to increase the average R$^2$ from 0.630 to 0.643. Likewise, we can also observe the contribution of the textual information from the results of DL-LSTM, where including this information allows the model to reduce the MAE from 103.9 to 100.4.

\begin{table}[t!]
\caption{Average results (and standard deviations) for the Barclays Center area over 30 executions for the proposed model (DL-FC), distinguishing between event and non-event days.}
\begin{center}
\small
\setlength\tabcolsep{5.2pt} 
\begin{tabular}{l|c|c|c|c}
& \multicolumn{2}{c|}{Non-event days} & \multicolumn{2}{c}{Event days}\\
Method & MAE & MAPE & MAE & MAPE\\
\hline
DL-LSTM L & 113.8 ($\pm$2.0) & 13.8 ($\pm$0.5) & 145.5 ($\pm$6.0) & 18.7 ($\pm$0.3) \\
DL-LSTM L+W & 119.7 ($\pm$2.6) & 15.0 ($\pm$0.4) & 141.5 ($\pm$1.6) & 18.7 ($\pm$0.1) \\
DL-LSTM L+W+E & 92.2 ($\pm$1.8) & 11.3 ($\pm$0.3) & 119.0 ($\pm$0.3) & 15.9 ($\pm$0.2) \\
DL-LSTM L+W+E+T & 85.3 ($\pm$1.4) & 10.2 ($\pm$0.1) & 118.7 ($\pm$1.5) & 16.0 ($\pm$0.3) \\
DL-FC L & 110.3 ($\pm$1.9) & 13.4 ($\pm$0.3) & 134.5 ($\pm$0.7) & 18.9 ($\pm$0.2) \\
DL-FC L+W & 109.0 ($\pm$1.9) & 13.2 ($\pm$0.4) & 134.2 ($\pm$0.9) & 18.7 ($\pm$0.2) \\
DL-FC L+W+E & 83.8 ($\pm$1.4) & 10.1 ($\pm$0.2) & 110.6 ($\pm$1.2) & 15.9 ($\pm$0.2) \\
DL-FC L+W+E+T & \textbf{81.8} ($\pm$0.8) & \textbf{9.8} ($\pm$0.1) & \textbf{108.4} ($\pm$1.1) & \textbf{15.5} ($\pm$0.2)
\end{tabular}
\end{center}
\label{table:barclaysb}
\end{table}%

In order to perform a more fine-grained analysis of the results of the proposed approach, Table~\ref{table:barclaysb} shows the MAE and MAPE for different day types: event days and non-event days. As the results show, the use of textual information about events is able to reduce the forecasting error for both types of day. Naturally, the error is larger in event days, since the latter correspond to non-recurrent scenarios, which are inherently harder to forecast. 

\begin{table}[t!]
\caption{Average results (and standard deviations) for the Terminal 5 area over 30 executions.}
\begin{center}
\small
\setlength\tabcolsep{5.2pt} 
\begin{tabular}{l|c|c|c|c}
Method & MAE & RMSE & MAPE & R$^2$ $(\times 100)$\\
\hline
SVR L & 186.7 ($\pm$0.0) & 252.2 ($\pm$0.0) & 20.4 ($\pm$0.0) & 43.7 ($\pm$0.0)\\
SVR L+W & 185.3 ($\pm$0.0) & 251.6 ($\pm$0.0) & 20.2 ($\pm$0.0) & 44.0 ($\pm$0.0)\\
SVR L+W+E & 177.0 ($\pm$0.0) & 244.4 ($\pm$0.0) & 19.1 ($\pm$0.0) & 47.1 ($\pm$0.0)\\
GP L & 204.5 ($\pm$0.0) & 264.6 ($\pm$0.0) & 22.5 ($\pm$0.0) & 38.0 ($\pm$0.0)\\
GP L+W & 204.2 ($\pm$0.0) & 264.8 ($\pm$0.0) & 22.5 ($\pm$0.0) & 37.9 ($\pm$0.0)\\
GP L+W+E & 184.4 ($\pm$0.0) & 250.3 ($\pm$0.0) & 20.3 ($\pm$0.0) & 44.5 ($\pm$0.0)\\
DL-LSTM L & 174.4 ($\pm$3.6) & 247.7 ($\pm$3.3) & 18.0 ($\pm$0.5) & 45.7 ($\pm$1.5)\\
DL-LSTM L+W & 172.9 ($\pm$3.4) & 245.5 ($\pm$2.3) & 17.8 ($\pm$0.5) & 46.6 ($\pm$1.0)\\
DL-LSTM L+W+E & 164.1 ($\pm$7.7) & 236.0 ($\pm$5.9) & 16.9 ($\pm$0.8) & 50.7 ($\pm$2.5)\\
DL-LSTM L+W+E+T & 160.8 ($\pm$6.1) & 233.5 ($\pm$4.9) & 16.7 ($\pm$0.7) & 51.7 ($\pm$2.0)\\
DL-FC L & 181.4 ($\pm$2.5) & 250.0 ($\pm$1.6) & 19.7 ($\pm$0.3) & 44.6 ($\pm$0.7)\\
DL-FC L+W & 180.7 ($\pm$2.1) & 250.1 ($\pm$1.3) & 19.6 ($\pm$0.3) & 44.6 ($\pm$0.6)\\
DL-FC L+W+E & 168.1 ($\pm$2.8) & 242.7 ($\pm$1.9) & 18.1 ($\pm$0.3) & 47.8 ($\pm$0.8)\\
DL-FC L+W+E+T & \textbf{152.6} ($\pm$2.4) & \textbf{232.1} ($\pm$2.5) & \textbf{16.1} ($\pm$0.3) & \textbf{52.3} ($\pm$1.1)\\
\end{tabular}
\end{center}
\label{table:terminal5}
\end{table}%

Moving on to the Terminal 5 study area, Table~\ref{table:terminal5} shows the obtained results. Contrarily to the results for the Barclays Center area, we can observe that, in this case, LSTMs are the best performing approach for modeling the time-series data alone, which can be verified by comparing the results of the different models that only use lagged observations (or \textit{lags}) as input (``L"). We find this result to be quite interesting. Although our two case studies are instances of the same demand prediction problem, the inherent characteristics of their areas makes certain types of deep learning architectures more suitable than others. In fact, this was a key motivation for developing the two data fusion approaches proposed in Section~\ref{subsec:nn}. 

Regarding the contributions of the different sources of information, the results of Table~\ref{table:terminal5} suggest identical conclusions as the ones for the Barclays Center area. We can observe that including weather information (``W") increases the prediction performance of all methods. Similarly, including information about the presence of events (``E") reduces the prediction error of all methods quite significantly. For example, the MAE of DL-LSTM is reduced from 172.9 to 164.1. In the case of DL-FC this reduction is even larger in magnitude: from 180.7 to 168.1. 

However, including information about the presence of events is not enough, since it only allows the deep learning approaches to model the average behavior corresponding to the mean increase in the number of taxi pickups when there are events taking place. Without delving into the information contained in the event descriptions, it is impossible for the models to find patterns that can help them discern impactful from low popularity events. By using word embeddings and convolutional layers, the proposed deep learning architectures are able to do so. The results from Table~\ref{table:terminal5} evidence this fact quite clearly. We can observe that, when combining text information (``T") with the remaining data sources, the proposed DL-FC is able to significantly reduce the prediction error. Namely, the MAE is reduced from 168.1 to 152.6, and the RMSE decreases from 242.7 to 232.1. Although in a smaller scale, a similar effect is also observed for the proposed DL-LSTM, in which case the MAE drops from 164.1 to 160.8. 

\begin{table}[t!]
\caption{Average results (and standard deviations) for the Terminal 5 area over 30 executions for the proposed model (DL-FC), distinguishing between event days and non-event days.}
\begin{center}
\small
\setlength\tabcolsep{5.2pt} 
\begin{tabular}{l|c|c|c|c}
& \multicolumn{2}{c|}{Non-event days} & \multicolumn{2}{c}{Event days}\\
Method & MAE & MAPE & MAE & MAPE\\
\hline
DL-LSTM L & 163.0 ($\pm$5.5) & 18.5 ($\pm$0.7) & 205.2 ($\pm$3.0) & 17.2 ($\pm$0.4) \\
DL-LSTM L+W & 159.9 ($\pm$5.9) & 18.0 ($\pm$0.8) & 206.5 ($\pm$5.0) & 17.0 ($\pm$0.4) \\
DL-LSTM L+W+E & 149.8 ($\pm$9.0) & 17.1 ($\pm$1.0) & 192.0 ($\pm$8.0) & 15.9 ($\pm$0.4) \\
DL-LSTM L+W+E+T & 148.9 ($\pm$7.8) & 17.0 ($\pm$0.9) & \textbf{188.4} ($\pm$5.8) & \textbf{15.6} ($\pm$0.4) \\
DL-FC L & 173.7 ($\pm$2.9) & 20.5 ($\pm$0.4) & 202.8 ($\pm$1.1) & 18.0 ($\pm$0.2) \\
DL-FC L+W & 173.3 ($\pm$2.4) & 20.4 ($\pm$0.3) & 200.3 ($\pm$1.7) & 17.8 ($\pm$0.2) \\
DL-FC L+W+E & 155.3 ($\pm$3.1) & 18.4 ($\pm$0.4) & 199.6 ($\pm$3.4) & 17.4 ($\pm$0.3) \\
DL-FC L+W+E+T & \textbf{138.1} ($\pm$3.2) & \textbf{16.1} ($\pm$0.5) & 190.2 ($\pm$2.8) & 16.2 ($\pm$0.2)
\end{tabular}
\end{center}
\label{table:terminal5b}
\end{table}%

As with the Barclays Center case study, we also compared the performance of the proposed deep learning approach in event days with non-event days. The obtained results are presented in Table~\ref{table:terminal5b}. The latter are very clear in showing that the use of textual information leads to a significant reduction in prediction error for both day types. From a data fusion perspective, these results clearly highlight how crucial data fusion of time-series data and textual information can be, in particular for the problem of predicting taxi demand in event areas considered in this paper. However, it should be noted that this is true because the textual data contains contextual information that correlates and explains some of the aspects of the time-series data. As explained earlier, this is a generally valid and broadly applicable assumption, that goes well beyond the transportation domain. 

Lastly, from the perspective of a transportation practitioners, it is important to note that by exploiting event information automatically extracted from the Web and by developing two novel deep learning architectures for combining this information with historical time-series data, were we able to reduce prediction error in event areas quite dramatically in both study areas. In the case of the Barclays Center area, we started with an initial MAPE of 15.6 for ``DL-FC L+W" and we were able to obtain a MAPE of 12.3 by using the full model ``DL-FC L+W+E+T". Likewise, in the case of the Terminal 5 area, we started with a MAPE of 19.6 for ``DL-FC L+W" and reached a MAPE of 16.1 for the full model ``DL-FC L+W+E+T". These are very significative improvements that emphasize the importance of accounting for the effect of special events when forecasting mobility demand in dynamic and lively urban areas. 

\section{Conclusion}
\label{sec:conclusion}

We proposed two deep learning data fusion architectures that combine temporal data with information in the form of unstructured text in order to improve time-series forecasts. The proposed approaches make use of word embeddings and convolutional layers to capture patterns in the text that correlate with the time-series observations, which are modeled either using LSTMs or a stack of fully-connected layers. The neural network architectures developed in this paper are then able to construct latent representations of the time-series and textual data, which can then be combined to produce more accurate forecasts. Although the cross-modal data fusion methodology that we propose is quite general, we applied it in the transportation domain, Specifically, we considered the problem of taxi demand forecasting in event areas. Using data from a large-scale publicly-available dataset of taxi trips in New York and event data automatically mined from the Web, we empirically demonstrate that fusing these two very different sources of data leads to significant reductions in forecasting error. Furthermore, our results show that by making use of event information from the Web, the proposed deep learning architectures are able to improve the quality of their predictions quite dramatically, thus significantly outperforming other popular time-series forecasting methods from the state of the art that do not account for event information. Hence, besides the value of data fusion of text and time-series data, our empirical results also highlight the need for accounting for the effect of events when modeling mobility demand. Lastly, we make all the code and datasets used in this paper publicly available, in an effort to set it as clear well-defined baseline for developing other cross-modal data fusion approaches. 

In future work, we aim at exploring how to make use of the data fusion techniques proposed in order to develop city-wide spatio-temporal deep learning models that account for information about all the events that take place across the city, including textual information from their titles and descriptions in order to help the models distinguish impactful from negligible events. 

\section*{Acknowledgments}

The research leading to these results has received funding from the People Programme (Marie Curie Actions) of the European Union's Seventh Framework Programme (FP7/2007-2013) under REA grant agreement no. 609405 (COFUNDPostdocDTU), and from the European Union?s Horizon 2020 research and innovation programme under the Marie Sklodowska-Curie Individual Fellowship H2020-MSCA-IF-2016, ID number 745673.

\section*{References}

\bibliography{deep-fusion-nyc-events_v2}

\end{document}